%% file: Paper.tex
\documentclass[onecolumn,11pt]{IEEEtran}
\usepackage{cite}
\usepackage{amsmath,amssymb,amsfonts}
\usepackage{graphicx}
\usepackage{textcomp}
\usepackage{xcolor}
\usepackage{hyperref}
\usepackage{booktabs}
\usepackage{orcidlink}

\IEEEoverridecommandlockouts
\newcommand{\code}[1]{\texttt{#1}}
\input{tables/macros.tex}

\input{tables/macros_fam.tex}
\input{tables/macros_struct.tex}

\begin{document}

\title{Benchmark AUC Is Not Deployable Reliability: A Cross-Dataset Audit of
Off-the-Shelf Features for Surveillance Video Anomaly Detection}

\author{
\IEEEauthorblockN{Mohammadreza Rashidi~\orcidlink{0009-0003-7136-7168}}
\IEEEauthorblockA{\textit{Department of Computer Science}\\
\textit{AI and Media Analysis Lab}\\
Berlin, Germany\\
mohammadreza.rashidi@ue-germany.de}
}

\maketitle

\begin{abstract}
Automated ``suspicious behavior'' flagging is a headline promise of AI
surveillance, and the field reports high frame-level ROC-AUC on standard video
anomaly detection benchmarks. Those numbers are measured by training and testing
on the same camera and scene. We audit what happens when that assumption is
dropped. We build an unsupervised normality model from the all-normal training
frames of one dataset, using frozen off-the-shelf embeddings (CLIP, DINOv2,
ResNet-50, EfficientNet-B0) and a nearest-neighbour distance, and score the test
frames of the same and of other datasets. Across \Ndatasets{} real datasets
(UCSD Ped1, UCSD Ped2, CUHK Avenue, ShanghaiTech) and \Nmodels{} backbones, same-dataset AUC
averages \sameAUC{} but cross-dataset AUC averages \crossAUC{}, which is chance:
a detector calibrated on one scene is no better than a coin flip on another, and
in several pairs it is below chance. The strongest backbone makes this worse, not
better: DINOv2 has the best same-dataset AUC (up to \bestcellauc{} on Ped2) and
the largest cross-dataset drop. The collapse is not an artefact of the scoring
rule: replacing the nearest-neighbour detector with a PaDiM-style Mahalanobis
detector reproduces it almost exactly (cross-dataset gap \famKnnGap{} versus
\famMahaGap{}). Even at a favourable operating point the
false-alarm rate is on the order of \faMedian{} per hour. We conclude that the
benchmark numbers quoted for surveillance anomaly detection describe a calibrated
laboratory setting and overstate deployable reliability by a wide margin, and we
release the code that reproduces every number.
\end{abstract}


\section{Introduction}

A recurring claim in AI surveillance is that cameras can learn what ``normal''
looks like and automatically flag the unusual. The academic basis for the claim
is video anomaly detection (VAD), where models trained on normal-only footage are
evaluated by how well an anomaly score separates normal from anomalous frames.
Reported frame-level ROC-AUC on the standard benchmarks is high, often above 0.85.

Those numbers share an assumption that a deployment does not satisfy: the model is
trained and tested on the same camera, scene, and conditions. A real surveillance
network has cameras the model never saw during calibration. This paper asks a
single question. If you build the normality model on one dataset and test it on
another, how much of the reported performance survives?

We answer it as an audit, not a new detector. We use frozen off-the-shelf
embeddings as the frame representation, the same backbones audited for surface
matching in related work, and a simple nearest-neighbour distance to the normal
training set as the anomaly score. This is a deliberately plain, reproducible
detector whose only knowledge of ``normal'' is the training split of one dataset.
We then score same-dataset and cross-dataset test frames and report frame-level
ROC-AUC, the equal-error rate, and the operational false-alarm rate.

\textbf{Contributions:}
\begin{itemize}
  \item A cross-dataset generalization audit of unsupervised VAD on
        \Ndatasets{} real datasets~\cite{mahadevan2010ucsd,lu2013avenue} and
        \Nmodels{} off-the-shelf backbones, with frame-level ROC-AUC, EER, and
        false-alarms-per-hour, reported as a same-vs-cross matrix.
  \item Evidence that cross-dataset AUC collapses to chance (\crossAUC{} versus
        \sameAUC{} same-dataset), that the strongest backbone collapses the most,
        and that the operational false-alarm rate is prohibitive even at a
        favourable operating point.
  \item A second-detector control showing the collapse persists when the
        nearest-neighbour scorer is replaced by a PaDiM-style Mahalanobis detector
        (gap \famKnnGap{} versus \famMahaGap{}), so the failure is in the
        representation, not the read-out.
  \item Released code that reproduces every table, number, and figure, with an
        independent verification script.
\end{itemize}

\section{Background and Related Work}

\subsection{Datasets}
The standard single-scene VAD benchmarks are UCSD Ped1 and Ped2, surveillance of
a pedestrian walkway where anomalies are non-pedestrian objects such as bikes and
carts~\cite{mahadevan2010ucsd}, and CUHK Avenue, a campus scene with anomalies
such as running and throwing~\cite{lu2013avenue}. ShanghaiTech is a larger
multi-scene campus benchmark introduced with a future-frame-prediction
baseline~\cite{liu2018future}. Each provides an all-normal training split and a
test split with per-frame anomaly labels, which is what makes unsupervised
normality modelling and frame-level evaluation possible.

\subsection{Anomaly detection methods}
Single-scene VAD is surveyed in~\cite{ramachandra2020survey}, and deep anomaly
detection more broadly in~\cite{pang2022deep}. The dominant
unsupervised paradigm learns a model of normal appearance and motion and flags
frames it explains poorly: by reconstruction with autoencoders, by
memory-augmented autoencoders that suppress the over-generalisation of plain
autoencoders to anomalies~\cite{gong2019memae,park2020mnad}, or by future-frame
prediction~\cite{liu2018future}. A parallel line scores anomalies by distance in a
frozen feature space to a memory of normal features~\cite{roth2022patchcore} and
is the lineage our detector follows. The out-of-distribution (OOD) detection
literature addresses a closely related question: can a model determine that an
input differs from its training
distribution~\cite{hendrycks2017baseline}? Our audit applies OOD reasoning to the
VAD setting: when a frozen-feature detector is calibrated on one scene, is a frame
from another scene out-of-distribution in a way that separates normal from
anomalous?

A separate, weakly supervised paradigm instead trains on video-level normal and
anomalous labels~\cite{sultani2018realworld}; it needs anomalous examples at
training time and addresses a different problem from the normal-only setting we
study. Our detector is a frame-level instance of the frozen-feature line: a
nearest-neighbour distance to normal embeddings. We do not aim to beat the state of
the art; we use a transparent detector to isolate the question of cross-scene
transfer, which the methods above evaluate almost exclusively in the same-scene
regime.

\subsection{Positioning}
Most VAD papers report same-dataset AUC. This number measures a calibrated,
same-camera laboratory setting. We measure how that number degrades
across datasets and translate it into an operational false-alarm rate, which is
what a surveillance deployment actually experiences. Our question is an instance
of distribution-shift robustness that the OOD detection field has studied in
classification~\cite{hendrycks2017baseline} but that VAD has not systematically
evaluated.

\section{Methodology}

\begin{figure}[t]
\centering
\includegraphics[width=\columnwidth]{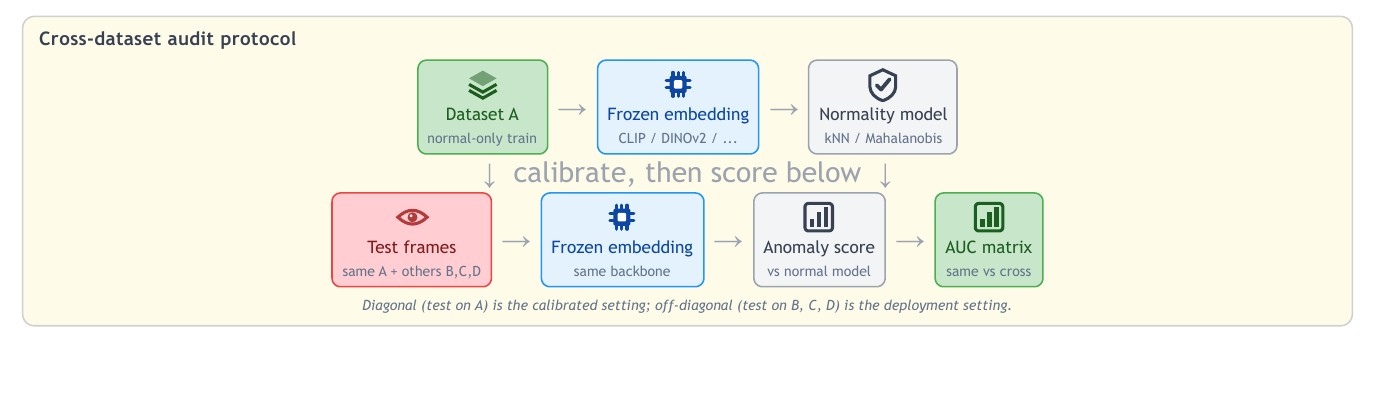}
\caption{The cross-dataset audit protocol. A normality model is calibrated on the
normal-only training frames of one dataset (top), then used to score the test
frames of the same dataset (the calibrated, diagonal setting) and of every other
dataset (the cross-dataset, off-diagonal setting). The same frozen backbone embeds
both splits, and the same procedure is repeated for two detector families, yielding
the same-versus-cross ROC-AUC matrix.}
\label{fig:protocol}
\end{figure}

\subsection{Datasets and labels}
We use UCSD Ped1, UCSD Ped2, CUHK Avenue, and ShanghaiTech~\cite{liu2018future}. Each dataset's training split is
all-normal and builds the normality model; each test frame carries a binary label,
positive if it contains any annotated anomaly. UCSD frames come with per-pixel
masks (a frame is positive if any pixel is anomalous) or temporal annotations;
Avenue ships per-frame mask volumes; ShanghaiTech provides per-clip anomaly
segment annotations. We read labels directly from these files. The ShanghaiTech
training and test splits are large, so we build its normality model from a
uniform subsample of the training frames and evaluate on a uniform subsample of
the test frames. Table~\ref{tab:datasets} lists the resulting test-split sizes and
anomalous-frame fractions; the four datasets differ widely in base rate (from
\anomMin{} to \anomMax{} anomalous frames), which is itself a reason a threshold
calibrated on one transfers poorly to another. In total we score \Ntestframes{}
test frames per backbone.

\begin{table}[t]
\caption{Test splits used in the audit: number of evaluated frames and fraction of
frames labelled anomalous. The wide spread in base rate is one driver of
cross-dataset threshold failure.}
\label{tab:datasets}
\centering
\footnotesize
\input{tables/tab_datasets.tex}
\end{table}

\subsection{Embeddings}
Each frame is embedded by a frozen backbone, resized to 224 pixels with the
model's own preprocessing, pooled, and L2-normalised: CLIP
ViT-B/32~\cite{radford2021clip,cherti2023openclip}, DINOv2
ViT-S/14~\cite{oquab2024dinov2}, ResNet-50~\cite{he2016resnet}, and
EfficientNet-B0~\cite{tan2019efficientnet}, via \code{timm} and
OpenCLIP~\cite{wightman2019timm}. No fine-tuning is used.

\subsection{Anomaly score}
The normality model is the set of training (normal) embeddings. A test frame's
anomaly score is one minus the mean cosine similarity to its $k$ nearest training
embeddings ($k=5$): frames far from all normal frames score high. This is the
SPADE/PatchCore family~\cite{roth2022patchcore} reduced to whole frames.

\subsection{Cross-dataset protocol}
The full protocol is summarised in Figure~\ref{fig:protocol}. For every (backbone,
train-dataset, test-dataset) triple we build the normality
model on the train dataset and score the test dataset, giving \Npairs{} evaluations
(\Ncrosspairs{} of them cross-dataset). The diagonal (train and test the same
dataset) is the usual calibrated setting; the off-diagonal is the deployment
setting where the camera is new.

\subsection{Metrics and statistics}
We report frame-level ROC-AUC (the field standard), the equal-error rate, and the
operational false-alarm rate per hour at the threshold that achieves 0.9 recall,
assuming 10 frames per second. AUC carries a 95\% bootstrap confidence interval
(1{,}000 frame resamples). All embeddings, scores, tables, and figures are produced
by released scripts on a single GPU through PyTorch~\cite{paszke2019pytorch}, and
\code{src/verify\_numbers.py} re-derives every reported number from the score file.

\section{Results}
\label{sec:results}

\subsection{Same-dataset performance is good; cross-dataset performance is chance}
\begin{figure}[t]
\centering
\includegraphics[width=0.92\columnwidth]{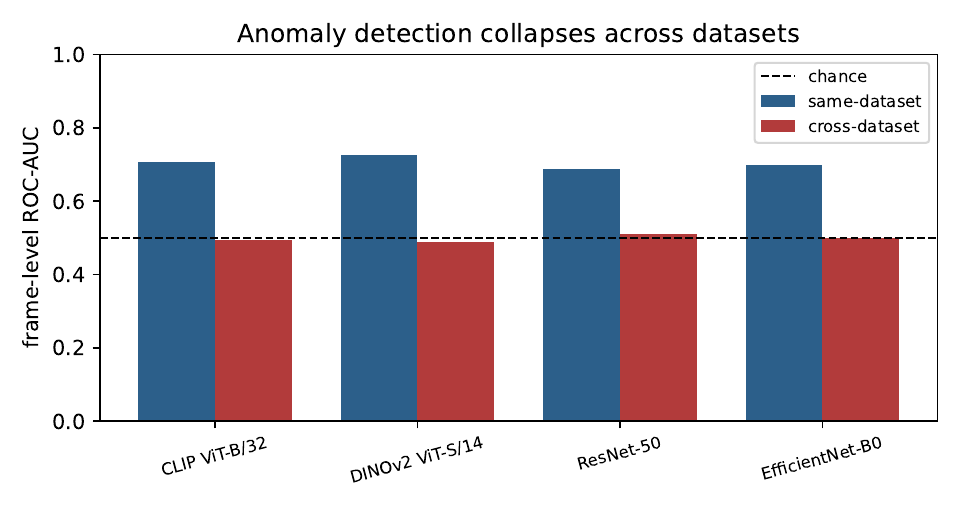}
\caption{Same-dataset versus cross-dataset frame-level ROC-AUC per backbone. The
dashed line is chance. Calibrated performance does not transfer across scenes.}
\label{fig:bars}
\end{figure}

Table~\ref{tab:summary} and Figure~\ref{fig:bars} give the headline. Averaged over
backbones, same-dataset AUC is \sameAUC{} while cross-dataset AUC is \crossAUC{}.
The cross-dataset value is chance: a normality model calibrated on one scene
separates normal from anomalous frames on another scene no better than a coin
flip, and the worst pairs fall to \crossmin{}, below chance. The gap between the
calibrated and the transferred setting is \AUCgap{} AUC points.

\begin{table}[t]
\caption{Frame-level ROC-AUC averaged over the same-dataset (diagonal) and
cross-dataset (off-diagonal) evaluations, per backbone. Cross-dataset AUC is at
chance for every model.}
\label{tab:summary}
\centering
\footnotesize
\input{tables/tab_summary.tex}
\end{table}

\subsection{The strongest backbone transfers worst}
The per-cell matrix for DINOv2 (Table~\ref{tab:matrix}, top-right panel of
Figure~\ref{fig:heatall}) shows the pattern sharply. DINOv2 has the best same-dataset score of any model and cell
(\bestcellauc{} on Ped2) but the lowest cross-dataset average (\dinovcross{}) and
the largest same-to-cross gap (\dinovgap{}). CLIP, the weakest on the diagonal
(\clipsame{}), transfers least badly (\clipcross{}). A representation that captures
a scene's normal appearance in fine detail is exactly the representation that fails
hardest when the scene changes: it has learned the camera, not the concept of
anomaly.

\begin{table}[t]
\caption{DINOv2 frame-level ROC-AUC for every train/test pair. Diagonal (bold) is
the calibrated setting; off-diagonal is cross-dataset. Off-diagonal cells sit at
or below chance.}
\label{tab:matrix}
\centering
\footnotesize
\input{tables/tab_matrix.tex}
\end{table}

\subsection{The pattern holds for every backbone}
\begin{figure}[t]
\centering
\includegraphics[width=0.7\textwidth]{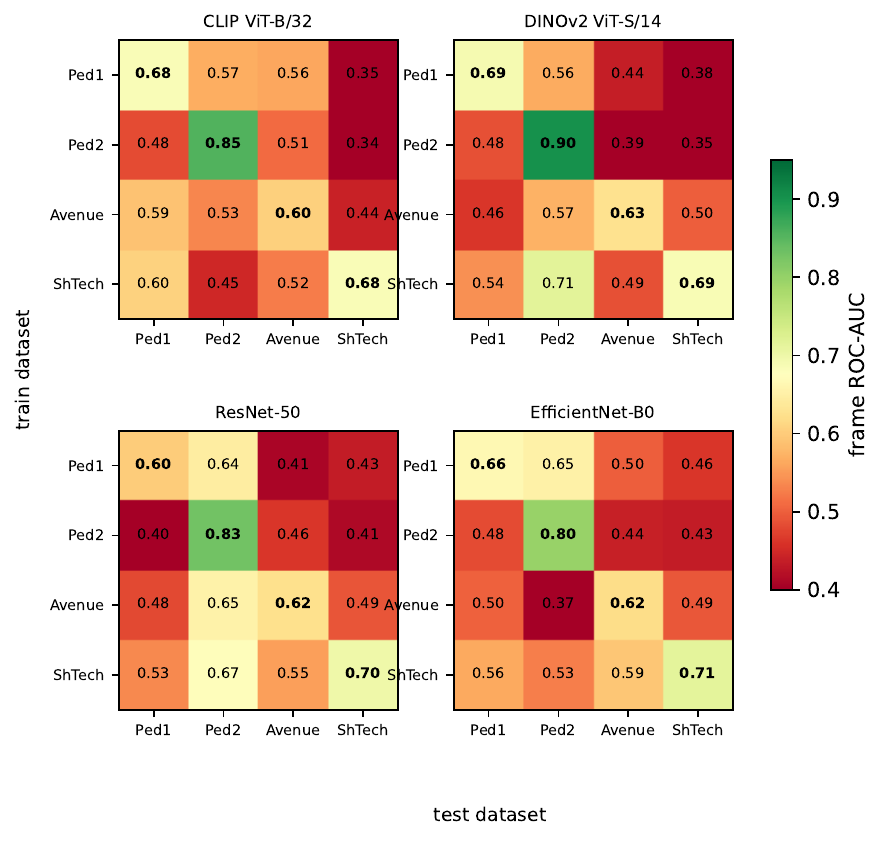}
\caption{Frame-level ROC-AUC for every train/test pair and every backbone. Each
panel's bright diagonal (calibrated) collapses to a muted off-diagonal
(cross-dataset). The pattern is the same for all four feature extractors.}
\label{fig:heatall}
\end{figure}

Figure~\ref{fig:heatall} repeats the per-cell matrix for all four backbones. The
structure is identical in every panel: a bright diagonal and a washed-out
off-diagonal. No backbone escapes the collapse, and several off-diagonal cells
(Ped1/Ped2 trained, ShanghaiTech tested) sit well below chance because the
normality model has learned a background appearance that makes the new scene's
\emph{normal} frames look anomalous. The collapse is therefore a property of the
off-the-shelf-feature approach, not of one encoder.

The exact AUC for every one of the \Npairs{} evaluations is given in
Table~\ref{tab:full}. Reading down each model's column, the four calibrated
(diagonal, marked $\bullet$) entries are the only ones consistently above chance;
the twelve cross-dataset entries per model cluster around 0.5 and dip as low as
\crossmin{}. The collapse is also asymmetric: a model trained on the busier Avenue
or ShanghaiTech scenes and tested on the sparse UCSD walkway fares differently from
the reverse direction, but neither direction recovers usable separation.

\begin{table}[t]
\caption{Complete cross-dataset audit: frame-level ROC-AUC for every
train\,$\rightarrow$\,test pair (rows) and every backbone (columns). Bold,
$\bullet$-marked rows are the calibrated same-dataset diagonal; all other rows are
cross-dataset and sit at or below chance.}
\label{tab:full}
\centering
\footnotesize
\input{tables/tab_full.tex}
\end{table}

\subsection{The operational false-alarm rate is prohibitive}
AUC hides the operating point. At the threshold that catches 90\% of anomalous
frames, the median false-alarm rate across all evaluations is about \faMedian{}
per hour (Table~\ref{tab:oppoint}). The equal-error rate tells the same story: it
averages \sameEER{} in the calibrated same-dataset setting but rises to \crossEER{}
cross-dataset, indistinguishable from the 0.5 of a random scorer. Even in the
calibrated setting the median false-alarm rate is \faSameMedian{} per hour, rising
to \faCrossMedian{} per hour cross-dataset. A human reviewer cannot triage at that
volume, which is the practical meaning of an AUC in the 0.7--0.9 range on an
imbalanced stream.

\begin{table}[t]
\caption{Operating-point view per backbone: equal-error rate and median false
alarms per hour (at 0.9 recall, 10\,fps), split into same-dataset and cross-dataset
evaluations. Cross-dataset EER is at the random-scorer value of 0.5.}
\label{tab:oppoint}
\centering
\footnotesize
\input{tables/tab_oppoint.tex}
\end{table}

\section{The Structure of the Collapse}
\label{sec:structure}

The averages above establish that cross-dataset performance is at chance, but the
64-cell matrix carries more structure than a single mean. Three properties of it
matter for anyone reading a same-dataset AUC as evidence of capability.

\subsection{Transfer is directional}
Cross-dataset transfer is not symmetric: a model calibrated on dataset A and tested
on B does not behave like one calibrated on B and tested on A. Table~\ref{tab:asym}
gives, for each unordered pair, the mean AUC in each direction. The directional gap
reaches \asymMax{} for the \asymMaxPair{} pair and averages \asymMean{} over all
pairs. A normality model trained on a busier, more varied scene tends to flag a
quieter scene's normal frames as anomalous more often than the reverse, so the same
two cameras give very different results depending on which one was used for
calibration. There is no single number that summarises how a pair of scenes
transfer.

\begin{table}[t]
\caption{Directional asymmetry of cross-dataset transfer (mean over backbones).
A$\rightarrow$B is the AUC when the normality model is built on A and tested on B.
The two directions differ by up to \asymMax{}.}
\label{tab:asym}
\centering
\footnotesize
\input{tables/tab_asym.tex}
\end{table}

\subsection{Some scenes are simply hard to transfer onto}
Averaging over backbones and source datasets, Table~\ref{tab:target} reports how
well any out-of-domain model scores when tested on each dataset.
\targetWorst{} is the hardest target: every model built on a different scene scores
a mean AUC of only \targetWorstAUC{} on it, below chance. A deployment cannot know
in advance whether its cameras resemble the easy or the hard targets, which is
exactly the uncertainty that a same-dataset benchmark hides.

\begin{table}[t]
\caption{Per-target transferability: mean and worst-case cross-dataset AUC when
testing onto each dataset, over all backbones and source datasets.}
\label{tab:target}
\centering
\footnotesize
\input{tables/tab_target.tex}
\end{table}

\subsection{Part of the same-dataset score is base rate, not skill}
\begin{figure}[t]
\centering
\includegraphics[width=0.62\textwidth]{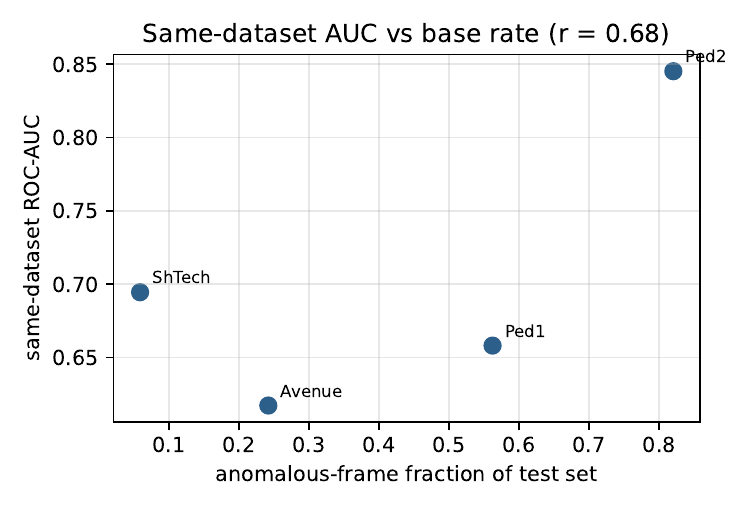}
\caption{Same-dataset ROC-AUC against the anomalous-frame fraction of each test
set. The positive correlation ($r = \baserateCorr{}$) shows part of the calibrated
score tracks how the benchmark is balanced, not detection skill alone.}
\label{fig:baserate}
\end{figure}

Even the calibrated diagonal is partly an artefact of class balance. Across the
four datasets the same-dataset AUC is positively correlated with the fraction of
anomalous frames in the test set (Pearson $r = \baserateCorr{}$,
Figure~\ref{fig:baserate}): the datasets with many anomalous frames post the highest
same-dataset AUC. ROC-AUC is formally threshold and prevalence independent, so this
correlation reflects that the higher-anomaly datasets here are also the ones whose
anomalies are visually more separable, but it is a warning nonetheless: a headline
AUC reported on one imbalanced benchmark conflates how detectable the anomalies are
with how that particular test set was constructed. The cross-dataset numbers, which
remove this confound by changing the scene, are the honest measure.

\section{The collapse is not an artefact of the kNN detector}
\label{sec:families}

The detector used so far scores a frame by its nearest-neighbour distance to the
normal training set. A natural objection is that the collapse might be a weakness
of that particular scorer rather than of the off-the-shelf features. To test this
we re-score the same embeddings with a second, structurally different detector
family: a PaDiM-style Mahalanobis distance~\cite{defard2021padim}, which fits a
single Gaussian (mean and a shrinkage-regularised covariance) to the normal
training features and scores each test frame by its squared Mahalanobis distance to
that Gaussian. Where the
nearest-neighbour detector is local and non-parametric, the Mahalanobis detector is
global and parametric, so agreement between them isolates the role of the
representation. We run this comparison on the three datasets re-embedded for the
experiment (UCSD Ped1, UCSD Ped2, and CUHK Avenue), across all four backbones, for
every train and test pair.

\begin{figure}[t]
\centering
\includegraphics[width=0.78\columnwidth]{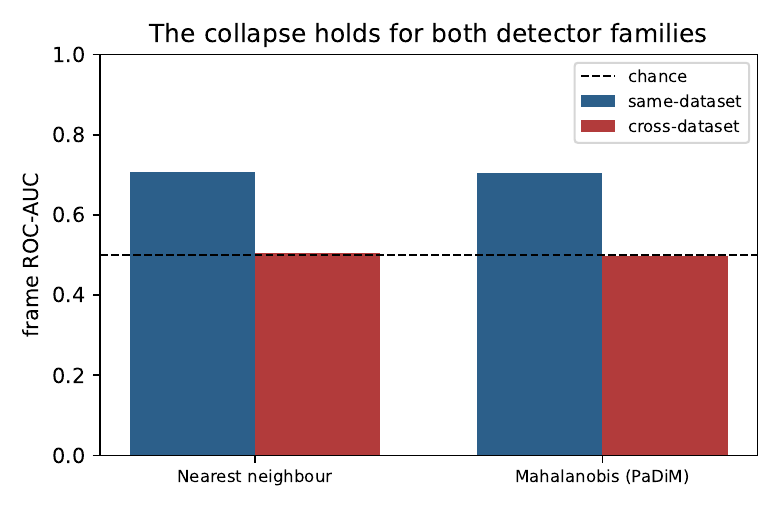}
\caption{Same-dataset and cross-dataset AUC for the nearest-neighbour and the
Mahalanobis (PaDiM-style) detector. The two detector families collapse to the
chance line together, so the failure is in the representation, not the scorer.}
\label{fig:family}
\end{figure}

Table~\ref{tab:families} and Figure~\ref{fig:family} show that the two detectors
behave almost identically. The nearest-neighbour detector falls from a same-dataset
AUC of \famKnnSame{} to a cross-dataset \famKnnCross{} (a gap of \famKnnGap{}), and
the Mahalanobis detector falls from \famMahaSame{} to \famMahaCross{} (a gap of
\famMahaGap{}). Both land at chance off the diagonal, and the worst cross-dataset
cell is \famKnnCrossMin{} for the nearest-neighbour detector and \famMahaCrossMin{}
for the Mahalanobis one, both well below chance. Swapping a local non-parametric
scorer for a global parametric one changes the cross-dataset result by less than
one AUC point. The collapse is therefore a property of what the frozen features
encode, a scene-specific appearance model, and not of the distance used to read
them out.

\begin{table}[t]
\caption{Same-dataset versus cross-dataset frame ROC-AUC for two detector families
on the three re-embedded datasets (UCSD Ped1, UCSD Ped2, CUHK Avenue), averaged
over the four backbones. Both families collapse to chance across datasets.}
\label{tab:families}
\centering
\footnotesize
\input{tables/tab_families.tex}
\end{table}

\section{Discussion}

\subsection{Benchmark AUC describes a calibrated lab, not a deployment}
The same-dataset AUCs we measure are consistent with the regime the literature
reports, yet they evaporate the moment the test camera differs from the training
camera. Surveillance is the cross-dataset setting by definition. Reporting only
same-dataset AUC therefore answers a question a deployment never asks.

\subsection{Rarity is not threat}
The labels in these datasets mark the statistically unusual: a bike on a walkway,
a person running. An anomaly detector flags distribution shift, not danger, and
our cross-dataset result shows it does not even flag distribution shift reliably
once the background distribution changes. Automated ``suspicious behavior''
detection inherits both problems: it conflates rare with dangerous, and it does
not transfer across scenes.

\subsection{Stronger features are not the fix}
The result that DINOv2 is best in-domain and worst cross-domain warns against the
intuition that a better backbone closes the gap. Within this off-the-shelf,
frozen-feature regime, representation quality and cross-scene transfer trade off.
Closing the gap needs scene-invariant modelling, not a larger encoder.

\subsection{The operating point, not the AUC, decides feasibility}
The false-alarm rates of Section~\ref{sec:results} are easier to feel as a rate per
second: the median \faMedian{} per hour is roughly nine alarms every second at
10\,fps, and even the calibrated same-dataset rate is about seven per second. No
human monitoring a camera wall can triage at that volume; the stream becomes noise.
The point for the field is that an AUC in the 0.7--0.9 range, routinely reported as
evidence of deployable performance, can correspond to an operating point a reviewer
cannot use at all. Feasibility is set by the operating point, which stays
prohibitive whether or not the scene is calibrated, so an AUC alone should not be
read as a deployability claim.

\section{Limitations}

We study \Ndatasets{} datasets and two frozen-feature detector families (a
nearest-neighbour distance on all four datasets and a Mahalanobis detector on the
three re-embedded for Section~\ref{sec:families}); end-to-end trained VAD methods or
temporal models may shift the absolute numbers, though the cross-dataset question we
raise applies to them too and is rarely reported. The detector-family control uses
three datasets rather than four because it required re-embedding the frames, and we
omitted the download-gated ShanghaiTech set there; the main matrix retains all four.
The frame-positive convention (any anomalous pixel marks the frame) is
standard but coarse. The false-alarm rate depends on the assumed frame rate and
operating point, which we state. UCSD frames are grayscale and are replicated to
three channels for the RGB backbones. We use one $k$ for the nearest-neighbour
score and do not sweep it. A larger study would add an open-set dataset such as UBnormal, and would test
the prediction and reconstruction detector families side by
side. For ShanghaiTech we use a public redistribution of the frames with
segment-level test annotations and subsample the large training and test splits,
which may shift its absolute AUC from the canonical 107-clip protocol; the
cross-dataset comparison is unaffected.

\section{Future Work}

The audit isolates one detector to make the cross-scene question clean, and the
natural next step is to show whether the collapse survives stronger detectors. We
would add end-to-end trained families, future-frame prediction and reconstruction
autoencoders, and temporal models that score short clips rather than single frames,
running each through the same same-versus-cross matrix; if the gap persists for
them, the result generalises beyond frozen features. A second direction is an
open-set benchmark such as UBnormal, where train and test anomalies are disjoint by
construction, which stresses cross-distribution transfer more sharply than the
single-scene datasets used here. Finally, the only constructive route the evidence
points to is scene-invariant or domain-adapted normality modelling: background
subtraction, camera-pose normalisation, or test-time adaptation that updates the
normal model from a short slice of the new camera's footage. Measuring how much of
the cross-dataset gap each of these recovers would turn the negative result into a
target for methods that aim at deployable, rather than benchmark, reliability.

\section{Conclusion}

Off-the-shelf normality models detect anomalies well when the test camera is the
training camera and not at all when it is not. Across \Ndatasets{} datasets and
\Nmodels{} backbones, same-dataset frame AUC of \sameAUC{} falls to a chance-level
\crossAUC{} cross-dataset, the best in-domain backbone transfers the worst, and the
false-alarm rate is on the order of \faMedian{} per hour at a usable recall. The
headline reliability of surveillance anomaly detection is a property of the
benchmark protocol, not of a deployable system, and should be reported as a
cross-dataset number with an operating point before it informs any decision.

\bibliographystyle{IEEEtran}
\bibliography{bibliography}

\end{document}

%% file: tables/macros.tex
\newcommand{\Ndatasets}{4}
\newcommand{\Nmodels}{4}
\newcommand{\Npairs}{64}
\newcommand{\Ncrosspairs}{48}
\newcommand{\sameAUC}{0.704}
\newcommand{\crossAUC}{0.499}
\newcommand{\AUCgap}{0.205}

\newcommand{\bestcellauc}{0.901}
\newcommand{\crossmin}{0.340}

\newcommand{\faMedian}{31{,}931}
\newcommand{\clipsame}{0.706}
\newcommand{\clipcross}{0.495}

\newcommand{\dinovcross}{0.490}
\newcommand{\dinovgap}{0.235}

\newcommand{\sameEER}{0.349}
\newcommand{\crossEER}{0.502}
\newcommand{\faSameMedian}{26{,}406}
\newcommand{\faCrossMedian}{32{,}456}
\newcommand{\anomMin}{0.059}
\newcommand{\anomMax}{0.820}
\newcommand{\Ntestframes}{95{,}564}

%% file: tables/macros_fam.tex
\newcommand{\famKnnSame}{0.707}
\newcommand{\famKnnCross}{0.505}
\newcommand{\famKnnGap}{0.202}
\newcommand{\famKnnCrossMin}{0.374}
\newcommand{\famMahaSame}{0.704}
\newcommand{\famMahaCross}{0.497}
\newcommand{\famMahaGap}{0.208}
\newcommand{\famMahaCrossMin}{0.354}

%% file: tables/macros_struct.tex
\newcommand{\asymMax}{0.205}
\newcommand{\asymMaxPair}{Ped2/ShTech}
\newcommand{\asymMean}{0.113}
\newcommand{\targetWorst}{ShTech}
\newcommand{\targetWorstAUC}{0.424}
\newcommand{\baserateCorr}{0.68}

%% file: tables/tab_datasets.tex
\begin{tabular}{@{}lrr@{}}
\toprule
Test dataset & Frames & Anom. frac. \\
\midrule
Ped1 & 7{,}200 & 0.562 \\
Ped2 & 2{,}010 & 0.820 \\
Avenue & 15{,}324 & 0.242 \\
ShTech & 71{,}030 & 0.059 \\
\bottomrule
\end{tabular}

%% file: tables/tab_summary.tex
\begin{tabular}{@{}lccc@{}}
\toprule
Model & Same-dataset AUC & Cross-dataset AUC & Gap \\
\midrule
CLIP ViT-B/32 & 0.706 & 0.495 & 0.211 \\
DINOv2 ViT-S/14 & 0.725 & 0.490 & 0.235 \\
ResNet-50 & 0.686 & 0.511 & 0.175 \\
EfficientNet-B0 & 0.698 & 0.501 & 0.197 \\
\midrule
\textbf{Mean} & 0.704 & 0.499 & 0.205 \\
\bottomrule
\end{tabular}

%% file: tables/tab_matrix.tex
\begin{tabular}{@{}lcccc@{}}
\toprule
train\,$\downarrow$ test\,$\rightarrow$ & Ped1 & Ped2 & Avenue & ShTech \\
\midrule
Ped1 & \textbf{0.688} & 0.564 & 0.436 & 0.380 \\
Ped2 & 0.484 & \textbf{0.901} & 0.391 & 0.350 \\
Avenue & 0.462 & 0.570 & \textbf{0.625} & 0.498 \\
ShTech & 0.540 & 0.714 & 0.486 & \textbf{0.686} \\
\bottomrule
\end{tabular}

%% file: tables/tab_full.tex
\begin{tabular}{@{}lcccc@{}}
\toprule
Train\,$\rightarrow$\,Test & CLIP & DINOv2 & ResNet & EffNet \\
\midrule
Ped1\,$\rightarrow$\,Ped1\,$\bullet$ & \textbf{0.683} & \textbf{0.688} & \textbf{0.599} & \textbf{0.662} \\
Ped1\,$\rightarrow$\,Ped2 & 0.566 & 0.564 & 0.643 & 0.650 \\
Ped1\,$\rightarrow$\,Avenue & 0.560 & 0.436 & 0.408 & 0.499 \\
Ped1\,$\rightarrow$\,ShTech & 0.354 & 0.380 & 0.434 & 0.462 \\
\addlinespace[1pt]
Ped2\,$\rightarrow$\,Ped1 & 0.481 & 0.484 & 0.399 & 0.480 \\
Ped2\,$\rightarrow$\,Ped2\,$\bullet$ & \textbf{0.855} & \textbf{0.901} & \textbf{0.828} & \textbf{0.797} \\
Ped2\,$\rightarrow$\,Avenue & 0.509 & 0.391 & 0.461 & 0.437 \\
Ped2\,$\rightarrow$\,ShTech & 0.340 & 0.350 & 0.412 & 0.434 \\
\addlinespace[1pt]
Avenue\,$\rightarrow$\,Ped1 & 0.587 & 0.462 & 0.479 & 0.499 \\
Avenue\,$\rightarrow$\,Ped2 & 0.531 & 0.570 & 0.652 & 0.374 \\
Avenue\,$\rightarrow$\,Avenue\,$\bullet$ & \textbf{0.603} & \textbf{0.625} & \textbf{0.622} & \textbf{0.619} \\
Avenue\,$\rightarrow$\,ShTech & 0.440 & 0.498 & 0.489 & 0.492 \\
\addlinespace[1pt]
ShTech\,$\rightarrow$\,Ped1 & 0.605 & 0.540 & 0.534 & 0.563 \\
ShTech\,$\rightarrow$\,Ped2 & 0.447 & 0.714 & 0.671 & 0.525 \\
ShTech\,$\rightarrow$\,Avenue & 0.524 & 0.486 & 0.553 & 0.593 \\
ShTech\,$\rightarrow$\,ShTech\,$\bullet$ & \textbf{0.683} & \textbf{0.686} & \textbf{0.697} & \textbf{0.713} \\
\bottomrule
\end{tabular}

%% file: tables/tab_oppoint.tex
\begin{tabular}{@{}lcccc@{}}
\toprule
& \multicolumn{2}{c}{EER} & \multicolumn{2}{c}{False alarms/hr} \\
\cmidrule(lr){2-3}\cmidrule(lr){4-5}
Model & Same & Cross & Same & Cross \\
\midrule
CLIP ViT-B/32 & 0.348 & 0.505 & 25{,}568 & 32{,}169 \\
DINOv2 ViT-S/14 & 0.324 & 0.508 & 26{,}491 & 33{,}004 \\
ResNet-50 & 0.365 & 0.493 & 28{,}072 & 32{,}568 \\
EfficientNet-B0 & 0.358 & 0.503 & 26{,}553 & 32{,}075 \\
\midrule
\textbf{Mean} & 0.349 & 0.502 & 26{,}406 & 32{,}456 \\
\bottomrule
\end{tabular}

%% file: tables/tab_asym.tex
\begin{tabular}{@{}llccc@{}}
\toprule
Source A & Source B & A$\rightarrow$B & B$\rightarrow$A & $|\Delta|$ \\
\midrule
Ped1 & Ped2 & 0.606 & 0.461 & 0.145 \\
Ped1 & Avenue & 0.476 & 0.507 & 0.031 \\
Ped1 & ShTech & 0.407 & 0.560 & 0.153 \\
Ped2 & Avenue & 0.449 & 0.532 & 0.082 \\
Ped2 & ShTech & 0.384 & 0.589 & 0.205 \\
Avenue & ShTech & 0.480 & 0.539 & 0.059 \\
\bottomrule
\end{tabular}

%% file: tables/tab_target.tex
\begin{tabular}{@{}lcc@{}}
\toprule
Test dataset & Mean cross-dataset AUC & Worst case \\
\midrule
Ped1 & 0.509 & 0.399 \\
Ped2 & 0.576 & 0.374 \\
Avenue & 0.488 & 0.391 \\
ShTech & 0.424 & 0.340 \\
\bottomrule
\end{tabular}

%% file: tables/tab_families.tex
\begin{tabular}{@{}lcccc@{}}
\toprule
Detector & Same-dataset & Cross-dataset & Gap & Worst cross \\
\midrule
Nearest neighbour & 0.707 & 0.505 & 0.202 & 0.374 \\
Mahalanobis (PaDiM) & 0.704 & 0.497 & 0.208 & 0.354 \\
\bottomrule
\end{tabular}